\documentclass{article}
\usepackage{ijcai20}

\usepackage{graphicx}
\usepackage{subfigure}
\usepackage{amsmath,amssymb,amsfonts}
\usepackage{amsthm}
\usepackage{multirow}
\usepackage[misc]{ifsym}
\usepackage{lipsum}

\usepackage[hidelinks]{hyperref}
\usepackage[utf8]{inputenc}
\usepackage[small]{caption}
\usepackage{times}
\usepackage{soul}
\usepackage{url}
\usepackage{booktabs}
\usepackage{algorithm}
\usepackage{algorithmic}

\title{A Spike Learning System for Event-driven Object Recognition}

\author{
Shibo Zhou\thanks{These authors contributed equally to this work.}$^1$
\and
Wei Wang\footnotemark[1]$^2$
\and
Xiaohua Li$^1$\And
Zhanpeng Jin$^2$
\affiliations
$^1$Binghamton University, State University of New York\\
\emails
\{zhou19, xli\}@binghamton.edu\\
$^2$University at Buffalo, State University of New York
\emails
\{wwang49, zjin\}@buffalo.edu
}

\begin{document}
\maketitle

\begin{abstract}
Event-driven sensors such as LiDAR and dynamic vision sensor (DVS) have found increased attention in high-resolution and high-speed applications. A lot of work has been conducted to enhance recognition accuracy. However, the essential topic of recognition delay or time efficiency is largely under-explored. In this paper, we present a spiking learning system that uses the spiking neural network (SNN) with a novel temporal coding for accurate and fast object recognition. The proposed temporal coding scheme maps each event's arrival time and data into SNN spike time so that asynchronously-arrived events are processed immediately without delay. 
The scheme is integrated nicely with the SNN's asynchronous processing capability to enhance time efficiency. A key advantage over existing systems is that the event accumulation time for each recognition task is determined automatically by the system rather than pre-set by the user. The system can finish recognition early without waiting for all the input events.
Extensive experiments were conducted over a list of 7 LiDAR and DVS datasets. The results demonstrated that the proposed system had state-of-the-art recognition accuracy while achieving remarkable time efficiency. Recognition delay was shown to reduce by $56.3\%$ to $91.7\%$ in various experiment settings over the popular KITTI dataset.
\end{abstract}


\section{Introduction}
\label{intro}

Object recognition, as a fundamental and key computer vision (CV) technique, has been substantially investigated over the decades. With the help of deep neural networks (DNNs), great success has been achieved regarding recognition accuracy. However, nearly all of the existing solutions work on digital images or videos that are captured by traditional cameras at a fixed rate, commonly 30 or 60 fps.  
Such traditional frame-based cameras encounter severe challenges in many highly demanding applications that require high speed, high accuracy, or high dynamics, such as autonomous driving, unmanned aerial vehicles (UAV), robotics, gesture recognition, etc. \cite{hwu2018adaptive}.
Low frame rate means low temporal resolution and motion blur for high-speed objects. High frame rate leads to a large amount of data with substantial redundancy and a heavy computational burden inappropriate for mobile platforms.

To tackle the above challenges, one of the alternative approaches is event-based sensing \cite{gallego2019event}. Typical examples include light detection and ranging (LiDAR) sensor, dynamic vision sensor (DVS), and radio detection and ranging (Radar) sensor. 
LiDAR is a reliable solution for high-speed and precise object detection in a wide view at long distance and has become essential for autonomous driving. DVS cameras have significant advantages over standard cameras with high dynamic range, less motion blur, and extremely small latency \cite{lichtsteiner2008128}. 

Traditional cameras capture images or videos in the form of frames. Event-based sensors capture images as asynchronous events. Events are created at different time instances and are recorded or transmitted asynchronously. Each event may have time resolution in the order of microseconds. Because events are sparse, the data amount is kept low even in wide-area 3D spatial sensing or high-speed temporal sensing.

Although some research has been made for object recognition and classification based on event sensors, asynchronous object recognition is mostly still an open objective \cite{gallego2019event}. Existing methods usually accumulate all the events within a pre-set collection time duration to construct an image frame for recognition. This synchronous processing approach is straight forward and can employ existing DNN methods conveniently. But it overlooks the temporal asynchronous nature of the events and suffers from recognition delays caused by waiting for events accumulation. Too long an accumulation duration leads to image blurring while too short an accumulation duration loses object details. The fixed accumulation duration is a limiting factor to recognition accuracy in practice. 

As an example, in \cite{liu2016combined}, the DVS sensor had an event rate from 10K to 300K events per second. For fast object tracking, a short fixed accumulation duration of 20 ms was adopted, which resulted in only $200$ to $6000$ events for each $240\times 180$ image frame. Obviously, this could hardly provide sufficient resolution for the subsequent CNN-based classifier. Novel methods that can recognize objects asynchronously with the accumulation duration optimized according to the nature of object recognition tasks are more desirable.

As another example, a typical $5$ Hz LiDAR sensor needs 0.2 seconds to collect all the events for a frame image. During this waiting period, a car travels at $120$ km/hour can run near $7$ meters. For timely object recognition or accident warning, methods that can process events asynchronously without waiting for the accumulation of the final events are more desirable. 

In this paper, we propose a new spike learning system that uses the spiking neural network (SNN) with a novel temporal coding to deal with specifically the task of asynchronous event-driven object recognition. It can reduce recognition delay and realize much better time efficiency. It can maintain competitive recognition accuracy as existing approaches. 

Major contributions of this paper are:

\begin{itemize}
    \item We design a new spike learning system that can exploit both the asynchronous arrival time of events and the asynchronous processing capability of neuron networks to reduce delay and optimize timing efficiency. The first ``asynchronous" means that events are processed immediately with a first-come-first-serve mode. The second ``asynchronous" means that the network can output recognition results without waiting for all neurons to finish their work. Integrating them together can significantly enhance time efficiency, computational efficiency, and energy efficiency.
    \item For the first time, recognition time efficiency is defined and evaluated extensively over a list of event-based datasets as one of the major objectives of object recognition. 
    \item We develop a novel temporal coding scheme that converts each event's asynchronous arrival time and data to SNN spike time. It makes it possible for the learning system to process events immediately without delay and to use the minimum number of events for timely recognition automatically. 
    \item We conduct extensive experiments over a list of $7$ event-based datasets such as KITTI \cite{Geiger2012CVPR} and DVS-CIFAR10 \cite{li2017cifar10}. Experiment results demonstrate that our system had a remarkable time efficiency with competitive recognition accuracy. Over the KITTI dataset, our system reduced recognition delay by $56.3\%$ to $91.7\%$ in various experiment settings. 
\end{itemize}



The rest of the paper is organized as follows. Section \ref{sec:related work} introduces the related work. Section \ref{sec:methods} provides the details of the proposed spiking learning system. Experiment datasets and results are given in Sections \ref{sec:datasets} and \ref{sec:evaluation}, respectively. We conclude the paper in Section \ref{sec:conclusion}.

\section{Related Work}
\label{sec:related work}


LiDAR uses active sensors that emit their own laser pulses for illumination and detects the reflected energy from the objects. Each reflected laser pulse is recorded as an event.
From the events, object detection and recognition can be carried out by various methods, either traditional feature extraction methods or deep learning methods. 
Behley et al. \cite{behley2013laser} proposed a hierarchical segmentation of the laser range data approach to realize object detection. Wang and Posner \cite{wang2015voting} applied a voting scheme to process LiDAR range data and reflectance values to enable 3D object detection. Gonzales et al. \cite{gonzalez2017board} explored the fusion of RGB and LiDAR-based depth maps. Tatoglu and Pochiraju \cite{tatoglu2012point} presented techniques to model the intensity of the laser reflection during LiDAR scanning to determine the diffusion and specular reflection properties of the scanned surface. Hernandez et al. \cite{hernandez2014lane} took advantage of the reflection of the laser beam to identify lane markings on the road surface. 
Asvadi et al. \cite{asvadi2017depthcn} introduced a convolutional neural network (CNN) to process 3D LiDAR point clouds and predict 2D bounding boxes at the proposal phase. An end-to-end fully convolutional network was used for a 2D point map projected from 3D-LiDAR data in \cite{li2016vehicle}. Kim and Ghosh \cite{kim2016robust} proposed a framework utilizing fast R-CNN to improve the detection of regions of interest and the subsequent identification of LiDAR data. Chen et al. \cite{chen2017multi} presented a top view projection of the LiDAR point clouds data and performed 3D object detection using a CNN-based fusion network. 




DVS, also called neuromorphic vision sensor or silicon retina, records the changing of pixel intensity at fine time resolution as events. DVS-based object recognition is still at an early stage. 
Lagorce et al. \cite{lagorce2016hots} utilized the spatial-temporal information from DVS to build features and proposed a hierarchical architecture for recognition. Liu et al. \cite{liu2016combined} combined gray-scale Active Pixel Sensor (APS) images and event frames for object detection. Chen \cite{chen2018pseudo} used APS images on a recurrent rolling CNN to produce pseudo-labels and then used them as targets for DVS data to do supervised learning with the tiny YOLO architecture. 

Built on the neuromorphic principle, SNN is considered a natural fit for neuromorphic vision sensors and asynchronous event-based sensors. SNN imitates biological neural networks by directly processing spike pulses information with biologically plausible neuronal models \cite{maass1997networks}\cite{ponulak2011introduction}.
Regular neural networks process information in a fully synchronized manner, which means every neuron in the network needs to be evaluated. Some SNNs, on the contrary, can work in asynchronous mode, where not all neurons are to be stimulated \cite{susi2018fns}. 
The attempts of applying SNN for neuromorphic applications include pattern generation and control in neuro-prosthetics systems \cite{ponulak2006resume}, obstacle recognition and avoidance\cite{ge2017spiking}, spatio- and spectro-temporal brain data mapping \cite{kasabov2014neucube}, etc.
Attempts were also made to use SNN for object detection and recognition, either over traditional frame-based image data \cite{cannici2019asynchronous,zhang2019tdsnn,lee2016training},
or over event-based LiDAR and DVS data \cite{zhou2020deepscnn}\cite{wu2019direct}. 

A class of SNNs was developed with temporal coding, where spiking time instead of spiking rate or spiking count or spiking probability was used to encoding neuron information. The SpikeProp algorithm \cite{bohte2002error} described the cost function in terms of the difference between desired and actual spike times. It is limited to learning a single spike. Supervised Hebbian learning \cite{legenstein2005can} and ReSuMe \cite{ponulak2010supervised} were primarily suitable for the training of single-layer networks only. 


As far as we know, all the existing SNN works over event-based sensor data need a pre-set time duration 
to accumulate events into frame-based images before recognition. How to break this limitation to develop SNNs with the full asynchronous operation is still an open problem. 


\section{A Spike Learning System}
\label{sec:methods}

\begin{figure}[t]
\centering
\includegraphics[width=1\linewidth]{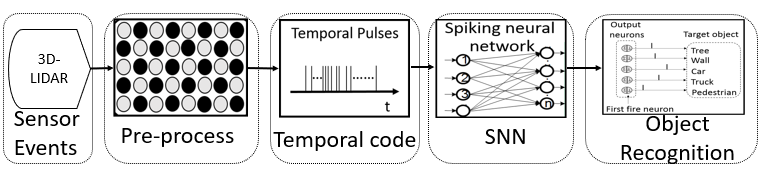}
\caption{Flow diagram of the proposed spike learning system.}
\label{fig:flowdiagram}
\end{figure}

Fig.~\ref{fig:flowdiagram} shows the workflow of our proposed spike learning system. The pipeline consists of three major blocks: 1) Pre-processing of asynchronous events from event-based sensors; 2) Temporal coding of the pre-processed events into SNN input spike time; 3) Object recognition with SNN.

\subsection{Pre-processing of Events}
The event data in standard LiDAR datasets are usually given as tuple $(x, y, z, r)$, where $(x, y, z)$ is the location of the object, and $r$ is the received light intensity. The events form a point cloud at a certain time-stamp. Existing LiDAR datasets usually contain this time-stamp only instead of event timing. 
For real-time applications, events may be collected asynchronously. Each event comes with its own arrival time $t_a$, which is the summation of laser pulse receiving time, LiDAR signal processing time, and data transmission time from the sensor to the learning system. 

With voxelization or other similar techniques \cite{zhou2020deepscnn},
we can compress the events data by quantizing the large spatial region into a small and fixed 3D integer grid $(x_v, y_v, z_v)$. For example, many papers quantize the KITTI dataset point cloud into a $768\times 1024 \times 21$ grid.
Let the spatial quantization step sizes in the three dimensions be $\Delta x$, $\Delta y$, and $\Delta z$, respectively. Then the event $(x, y, z, r)$ falls into the voxel
\begin{eqnarray}
    & {\cal V}(x_v, y_v, z_v) = \{(x, y, z): x_v \Delta_x \leq x < (x_v+1)\Delta_x, \nonumber \\
    & y_v \Delta_y \leq y < (y_v+1)\Delta_y, z_v \Delta_z \leq z < (z_v+1)\Delta_z\}.  
\end{eqnarray}
A voxel may have multiple or zero events due to events sparsity.
Its value can be set as the number of falling events, light intensity $r$, object distance $\sqrt{x^2+y^2+z^2}$, or laser light flying time $2\sqrt{x^2+y^2+z^2}/c$ with light speed $c$ \cite{zhou2020deepscnn}. In our experiments, we set the voxel value as 
\begin{equation}
 D(x_v, y_v, z_v) = \left\{ \begin{array}{ll} \frac{2\sqrt{x^2+y^2+z^2}}{c}, & (x, y, z) \in {\cal V}(x_v, y_v, z_v) \\
 0, & {\rm otherwise}
 \end{array} \right.
\end{equation}
We use the first arriving event $(x, y, z, r)$ inside this voxel to calculate $D(x_v, y_v, z_v)$. If no events falling inside this voxel, then $D(x_v, y_v, z_v)=0$. 

For DVS cameras, each event is recorded as $(x, y, t, p)$, where $(x, y)$ is the pixel coordinate in 2D space, $t$ is the time-stamp or arrival time of the event, and $p$ is the polarity indicating the brightness change over the previous time-stamp. The polarity is usually set as $p(x, y, t) = \pm 1$ or $p(x, y, t) = \{0, 1\}$ \cite{chen2019multi}. Pixels without significant intensity change will not output events. DVS sensors are noisy because of coarse quantization, inherent shot noise in photos, transistor circuit noise, arrival timing jitter, etc.

By accumulating the event stream over an exposure time duration, we can obtain an image frame. Specifically, accumulating events over exposure time from $t_0$ to $t_K$ gives the image
\begin{equation}
    D(x_v, y_v) = \sum_{t=t_0}^{t_K} p(x_v, y_v, t) + I(x_v, y_v),   \label{eq3.5}
\end{equation}
where $(x_v, y_v)$ is the pixel location, and $I(x_v, y_v)$ is the initial image at time $t_0$. We can set $I(x_v, y_v)=0$ from the start. Obviously, longer exposure duration $t_K-t_0$ leads to better image quality for slow-moving objects but blurring for fast-moving objects. Most existing methods pre-set an exposure duration, such as $100$ milliseconds for DVS-CIFAR10, to construct the image $D(x_v, y_v)$ for the subsequent recognition. In contrast, our proposed system does not have such a hard exposure time limitation and can automatically give recognition outputs within the best exposure time duration for the tasks.

\subsection{Temporal Coding for Spiking Neural Networks}
\label{sec:snn_model}

\begin{figure}[t]
\centering
\includegraphics[width=0.8\linewidth]{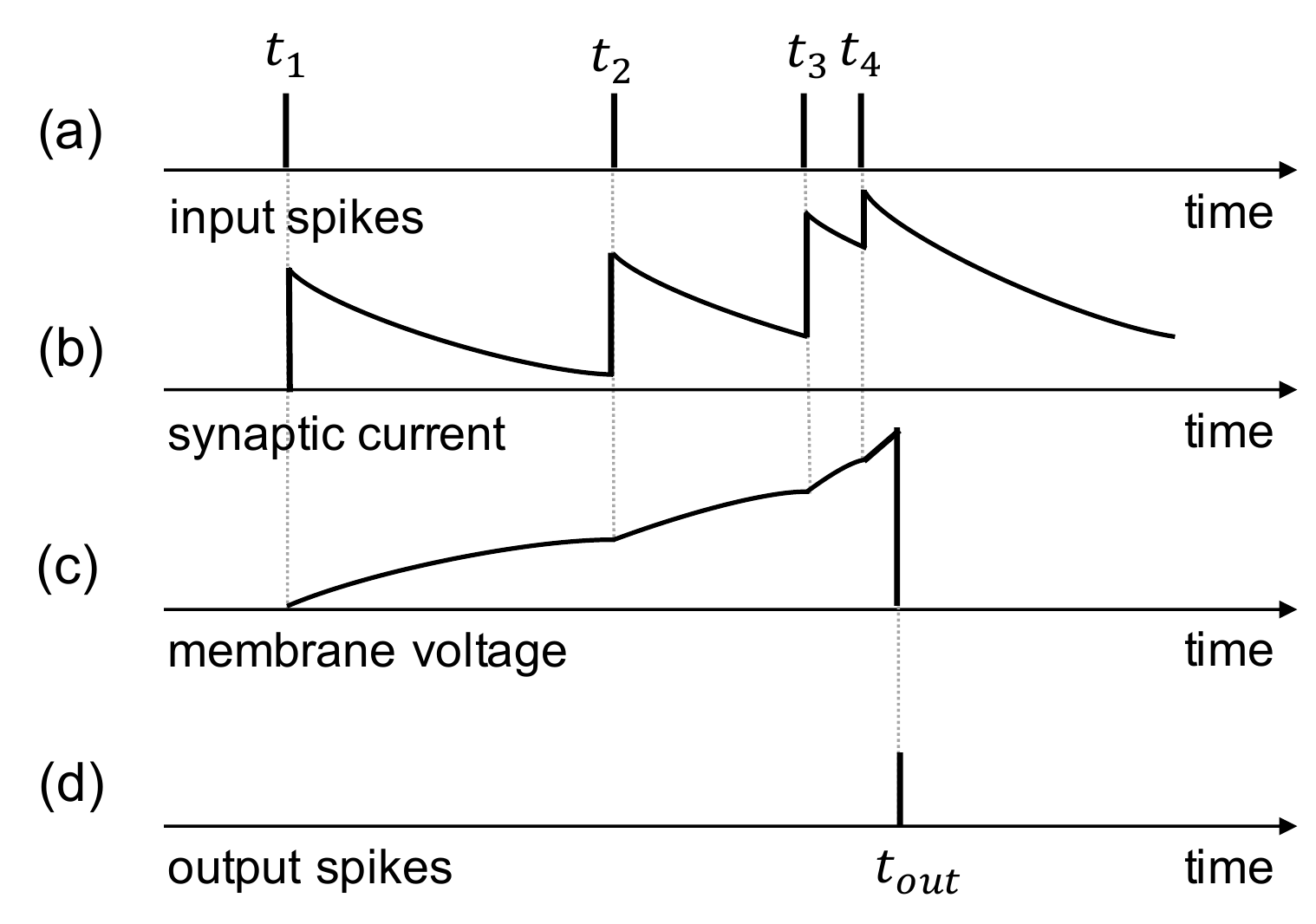}
\caption{The working principle of the n-LIF neuron model. (a) Four input neurons spike at time $t_i$, $i=1, \cdots, 4$. (b) Synaptic current $\kappa (t-t_i)$ jumps and decays over time. (c) Membrane voltage potential $v_j(t)$ rises towards the firing threshold. (d) The neuron $j$ emits a spike at time $t_j=t_{out}$ when the threshold is crossed.}
\label{fig:vmem}
\end{figure}

In SNNs, neurons communicate with spikes or action potentials through layers of the network. When a neuron's membrane potential reaches its firing threshold, the neuron will emit a spike and transmit it to other connected neurons \cite{ponulak2011introduction}. We adopt the spike-time-based spiking neuron model of \cite{mostafa2018supervised}.
Specifically, we use the non-leaky integrate-and-fire (n-LIF) neuron with exponentially decaying synaptic current kernels. The membrane potential is described by
\begin{equation}
\frac{dv_{j}(t)}{dt} = \sum_{i} w_{ji} \kappa(t-t_{i}), \label{eq3.10}
\end{equation}
where $v_{j}(t)$ is the membrane potential of neuron $j$, $w_{ji}$ is the weight of the synaptic connection from the input neuron $i$ to the neuron $j$, $t_i$ is the spiking time of the neuron $i$, and $\kappa(t)$ is the synaptic current kernel function. The value of neuron $i$ is encoded in the spike time $t_i$. The synaptic current kernel function determines how the spike stimulation decays over time. We use exponential decaying as given below
\begin{equation}
\kappa(t)=u(t)e^{-\frac{t}{\tau}},
\end{equation}
where $\tau$ is the decaying time constant, and $u(t)$ is the unit step function defined as
\begin{equation}
u(t) = \left\{ \begin{array}{ll}
1, \;\;\; & \text{if $t\geq0$}\\
0, \;\;\; & \text{otherwise}
\end{array} \right. .
\end{equation}
Fig. \ref{fig:vmem} illustrates how this neuron model works. A neuron is only allowed to spike once unless the network is reset or a new input pattern is presented. 

An analog circuit to implement this neuron was designed by \cite{zhou2020deepscnn} and was shown to be highly energy efficient. For training or digital (software) implementations, however, we do not need to emulate the operation (\ref{eq3.10}). Instead, we skip the dynamic time-evolution and consider only the member voltage at spiking time $t_{j}$. For this purpose, solving (\ref{eq3.10}) we get
\begin{equation}
    v_j(t_j) = \sum_{i\in C} w_{ji} {\tau} \left( 1- e^{-\frac{t_{j}-t_i}{\tau}} \right),
\end{equation}
where the set $C=\{i: t_i < t_{j}\}$ includes all (and only those) input neurons that spike before $t_{j}$. Larger $\tau$ leads to lower $v_j(t_j)$. For any $\tau$, we can find an appropriate voltage threshold $v_j(t_j)$ so that the activate input neuron set $C$ and the output spike time $t_j$ do not change. Therefore, in digital implementation, we can simply set both the voltage threshold and $\tau$ to 1. With $v_j(t_j)=1$, the neuron $j$'s spike time satisfies
\begin{equation}
e^{t_{j}} = \sum_{i\in C} e^{t_{i}} \frac{w_{ji}} {\sum_{\ell \in C} w_{j\ell}-1}. \label{eq3.20}
\end{equation}
In software SNN implementation, we can use directly $e^{t_i}$ as neuron value, calculate $w_{ji}/({\sum_{\ell \in C} w_{j\ell}-1})$ as weights, and (\ref{eq3.20}) is then the input-output equation of a feed-forward fully connected neural network layer.  We do not need other nonlinear activations because the weights are themselves nonlinear. 


At the first (or input) layer, we need to encode the pre-processed event data $D(x_v, y_v, z_v)$ into spike time $t_i$. Existing methods such as \cite{zhou2020deepscnn} simply let $t_i = D(x_v, y_v, z_v)$, which when applied onto the event-driven data will ignore the inherent temporal information and the asynchronous property of the events. To fully exploit the asynchronous events property, we propose the following new temporal coding scheme that encodes both event value $D(x_v, y_v, z_v)$ and the arrival time $t_a$ of each event.

Consider a LiDAR event $(x, y, z, r)$ arriving at time $t_a$. During preprocessing, assume it is used to update the voxel value $D(x_v, y_v, z_v)$. Also assume that this voxel $(x_v, y_v, z_v)$ corresponds to the $i$th input neuron in the SNN input layer. This neuron's spiking time is then set as
\begin{equation}
    t_i = \max\{\beta, t_a\} + \alpha D(x_v, y_v, z_v).   \label{eq3.25}
\end{equation}
where $\beta$ is a time parameter used to adjust delayed processing of early arrival events, and $\alpha$ is a constant to balance the value of $D(x_v, y_v, z_v)$ and $t_a$ so that the two terms in $t_i$ are not orders-of-magnitude different. 

The first term in the right of (\ref{eq3.25}) encodes the event arrival time. If $\beta = 0$, then there is no delay in encoding the arrival time: all the events are processed immediately upon arrival. If $\beta$ is set to the image frame time, we have the conventional synchronous event processing scheme where object recognition does not start until all the events are accumulated together. We can use $\beta$ to control the exposure time $t_K$. The second term encodes the event value. If $\alpha$ is set as $0$, then only the event arrival time is encoded. In this case, $\beta$ should be set as a small value, so that the temporal information could be fully exploited (e.g. if $\beta = 0$, then $t_i = t_a$). As a matter of fact, (\ref{eq3.25}) is a general temporal encoding framework. Various encoding strategies can be realized with appropriate parameters $\alpha$ and $\beta$, such as encoding event value only, encoding event arrival time only, and encoding both event value and arrival time.

For DVS sensors, assume similarly that the pixel $(x_v, y_v)$ is the $i$th input neuron. During the exposure time, while events are being accumulated into $D(x_v, y_v)$ according to (\ref{eq3.5}), the spiking time is set as the smallest $t_i$ that satisfies
\begin{equation}
     \sum_{t=t_0}^{t_i} p(x_v, y_v, t) + I(x_v, y_v) \geq \Gamma(t_i).      \label{eq3.26}
\end{equation}
$\Gamma(t)$ is a threshold function that can be set as a constant $\alpha$ or a linear decreasing function
\begin{equation}
    \Gamma(t) = \beta(t_K-t),
\end{equation}
with rate $\beta$. If $\beta=0$, then the neuron spikes immediately when the pixel value is positive. A sufficiently large $\beta$ effectively makes us wait and accumulate all the events to form a frame image before SNN processing. In this case, we fall back to the traditional synchronous operation mode.
If a pixel's intensity accumulates faster, then it spikes earlier. If the pixel accumulates slower or even stays near 0, then it may never spike. 

With the proposed temporal coding scheme, the system would be able to output a recognition decision asynchronously after some accumulation time $t$ between $t_0$ and $t_K$, and the event accumulation operation stops at $t$. Only the events during $t_0$ and $t$ are used for inference. Therefore, $t_0$ and $t_K$ can be simply set as the start and end time of the recognition task, such as an image frame time. This avoids the headache of looking for the best pre-set accumulation time. Note that no pre-set accumulation time can be optimal for all images. Some images need longer accumulation, while some other images need short accumulation. The proposed temporal coding enables our system to resolve this challenge in a unique way: We just pre-set a very large accumulation time $t_K$, and the system can automatically find the optimal accumulation time $t$ used for each image. In other words, instead of the fixed pre-set accumulation time, a unique event accumulation time $t$ (well before $t_K$) can be found for each image automatically by the system.

\subsection{Object recognition with SNN}


Many other SNNs use more complex neuron models or use spike counts or rates as neuron values. In comparison, our neuron model is relatively simple and has only a single spike, which makes our SNN easier to train and more energy-efficient when implemented in hardware. 
Based on the neuron input/output expression (\ref{eq3.20}), the gradient calculation and gradient-descent-based training become nothing different from conventional DNNs. 

\begin{figure}[t]
\centering
\includegraphics[width=\linewidth]{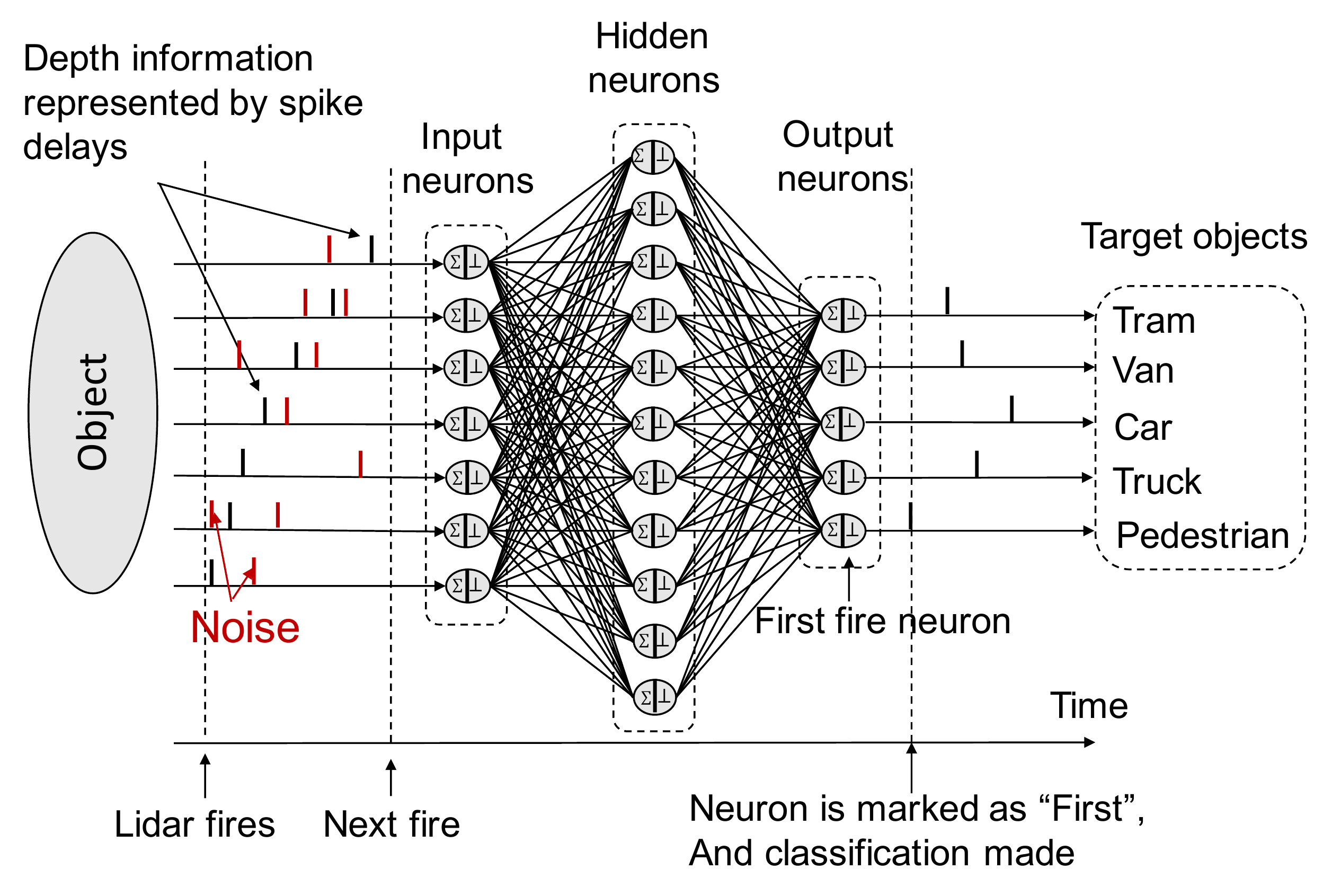}
\caption{SNN with spiking fully-connected (FC) layers for object recognition.}
\label{fig:implementation}
\end{figure}

\begin{figure}[t]
\centering
\includegraphics[width=1\linewidth]{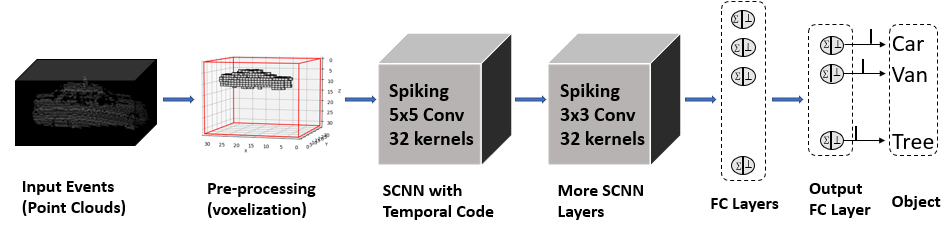}
\caption{SNN with multiple SCNN layers and FC layers for object recognition.}
\label{fig:scnn}
\end{figure}

Based on (\ref{eq3.20}), we can implement both spiking fully-connected (FC) layers and spiking convolutional neural network (SCNN) layers just as conventional DNN FC and CNN layers. Fig. \ref{fig:implementation} shows an SNN with two FC layers: one input layer with our temporal coding, one hidden FC layer, and one output FC layer. SCNN layers work in a similar way as traditional CNN but are equipped with spiking kernels. For more complex tasks, we can apply multiple SCNN layers and FC layers, as shown in Fig. \ref{fig:scnn}. Pooling layers such as max-pooling and average-pooling can also be used, which is the same as conventional DNNs. The standard back-propagation technique can be used to train the weights $w_{ji}$. 

For real-time object recognition, the SNN spiking time range is on the same scale as the event arrival time, specifically, starting at $t_0$ and ending at $t_K$. The SNN takes in spikes sequentially according to their arrival time.  Each neuron keeps accumulating the weighted value and comparing it with the threshold until the accumulation of a set of spikes can fire the neuron. Once the neuron spikes, it would not process any further input spikes unless reset or presented with a new input pattern. The recognition result is made at the time of the first spike among output neurons. Smaller $t_j$ or $e^{t_j}$ means stronger classification output. Also, smaller $t_j$ as output means that inference delay can be reduced, which is an outcome of the asynchronous working principle of SNN. 
 
Define the input of SNN as ${\bf z}_0$ with elements $z_{0,i} = e^{t_i}$, and the output of SNN as ${\bf z}_{L}$ with elements $z_{L, i}=e^{t_{L, i}}$. Then we have ${\bf z}_{L} = f({\bf z}_0; {\bf w})$ with nonlinear mapping $f$ and trainable weight ${\bf w}$ which includes all SNN weights $w_{ji}$ and the temporal coding parameter $\beta$. Let the targeting output be class $c$, then we train the network with the loss function

\begin{equation}
    {\cal L} ({\bf z}_L, c) = -\ln \frac{z_{L, c}^{-1}}{\sum_{i\neq c} z_{L, i}^{-1}} + k \sum_j \max\left\{0, 1-\sum_i w_{ji}\right\},   \label{eq3.50}
\end{equation}
where the first term is to make $z_{L, c}$ the smallest (equivalently $t_{L, c}$ the smallest)  one, while the second term is to make sure that the sum of input weights of each neuron be larger than $1$. The parameter $k$ adjusts the weighting between these two terms.
 

The training of (\ref{eq3.50}) can be conducted with the standard backpropagation algorithm similar to conventional DNNs, just as \cite{mostafa2018supervised}. Nevertheless, a problem of \cite{mostafa2018supervised} is that its presented algorithm did not require $t_j> t_i$ for $i \in {\cal C}$. This led to $t_j \leq t_i$ or even negative $t_j$ that was not practical. We corrected this problem and implemented the training of (\ref{eq3.50}) in the standard deep learning platform Tensorflow.

\section{Evaluation Datasets}
\label{sec:datasets}

\begin{figure}[t]
\centering
\frame{\includegraphics[width=0.9\linewidth]{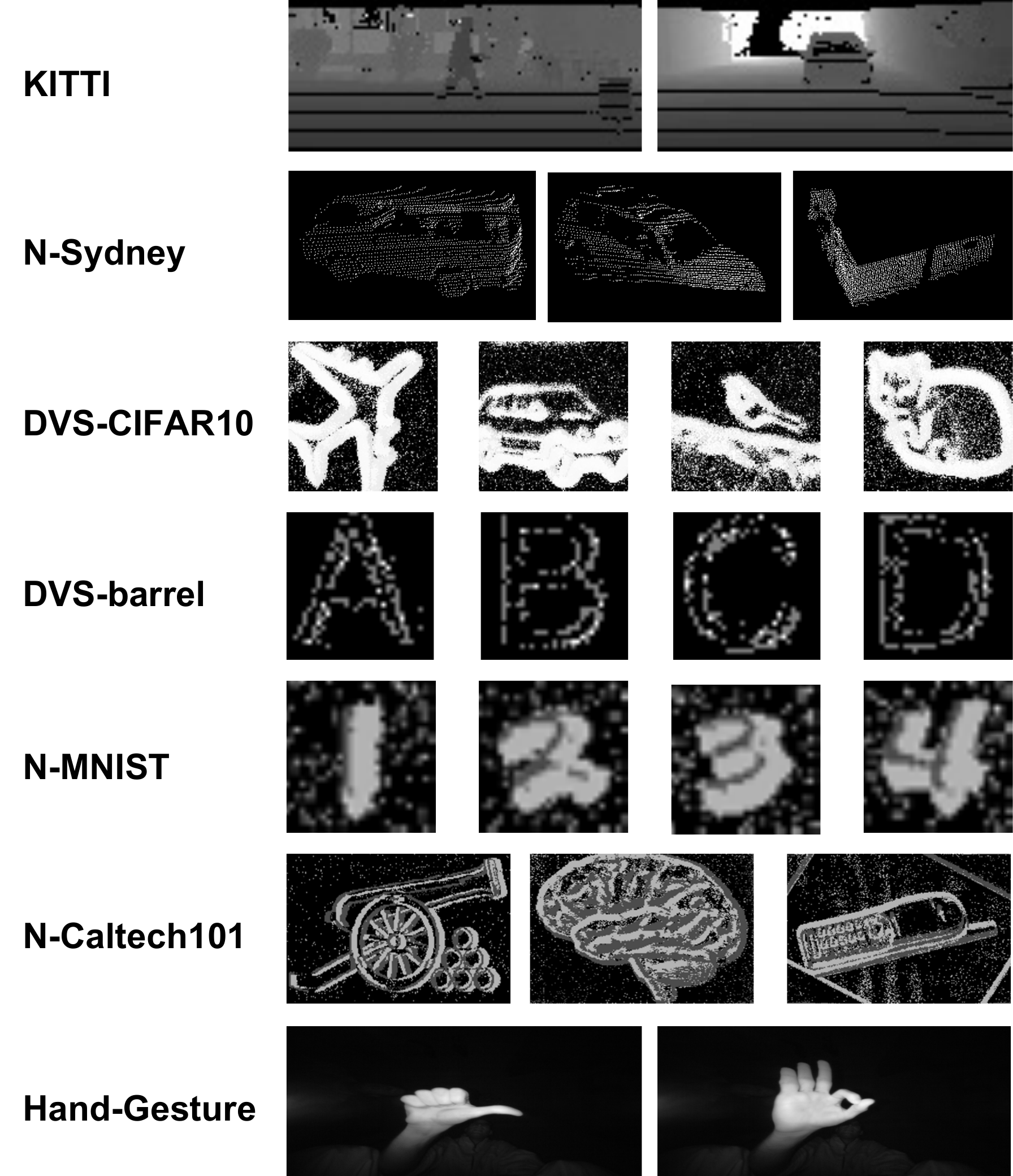}}
\caption{Sample images of LiDAR and DVS datasets used in this paper.}
\label{fig:imagesamples}
\end{figure}

To investigate the effectiveness of our proposed system, we evaluated it on a list of $7$ LiDAR and DVS datasets introduced below. Their sample images are shown in Fig. \ref{fig:imagesamples}.

\subsection{LiDAR Datasets}

\paragraph{\bf KITTI Dataset}
In order to evaluate the ability of our proposed system on complex real-life data in the autonomous driving scenario, we trained and tested it on the KITTI dataset \cite{Geiger2012CVPR}. We utilized the KITTI 3D object detection benchmark, specifically the point clouds data collected by Velodyne HDL-64E rotating 3D laser scanner, which provided 7481 labeled samples. However, the provided point clouds data can not be directly used because all label annotations (location, dimensions, observation angle, etc.) are provided in camera coordinates instead of Velodyne coordinates. 

\begin{figure}[tbp!]
\centering
\includegraphics[width=0.6\linewidth]{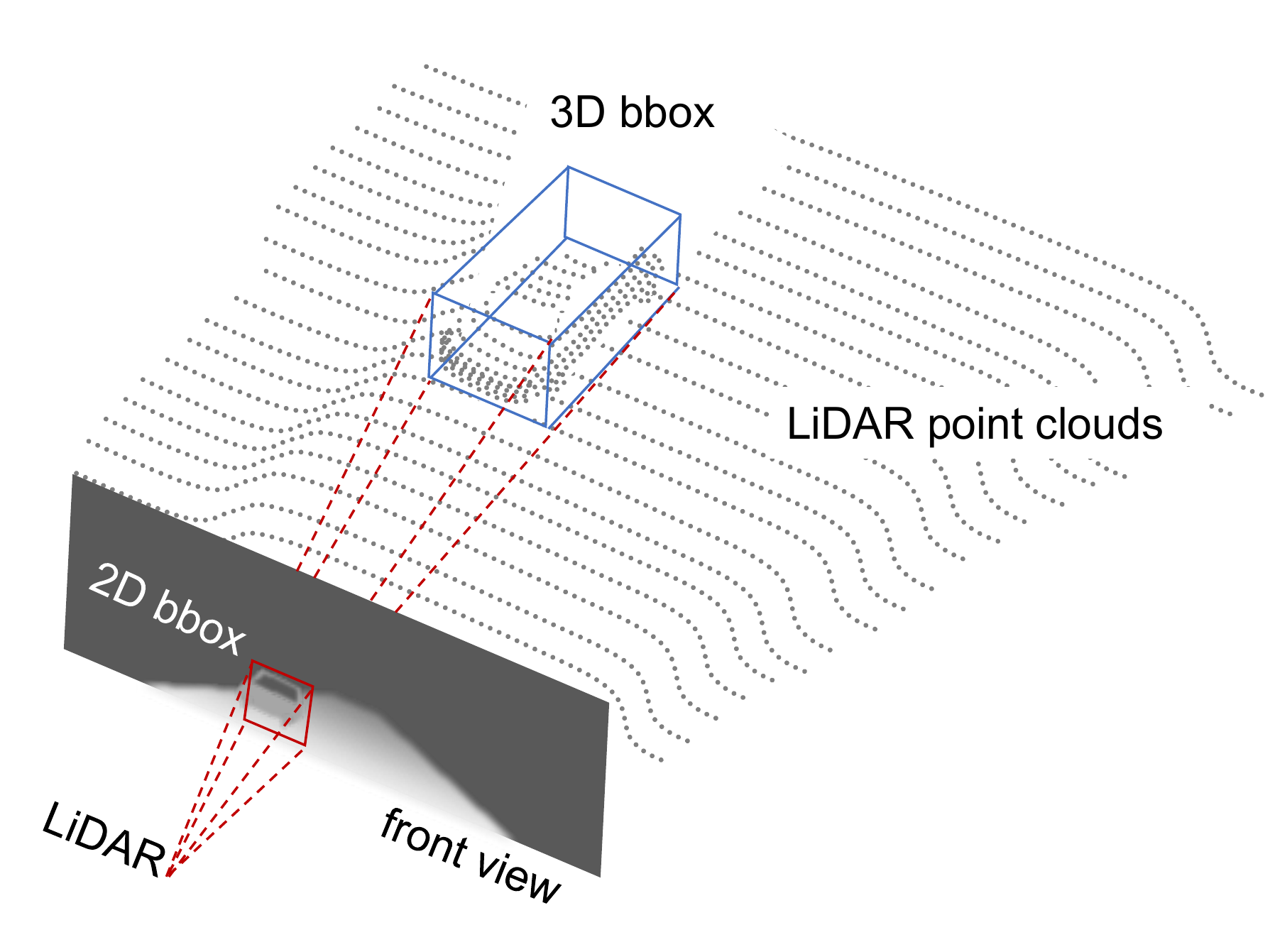}
\caption{Transformation of KITTI 3D Point Clouds into 2D LiDAR front view images.}
\label{fig:kitti_process}
\end{figure}

To convert point clouds data onto Velodyne coordinates, we first mapped point clouds data $(x, y, z)$ to an expanded front view $(x_{\rm front}, y_{\rm front})$, whose size was determined by the resolution of the LiDAR sensor. We used the transformation
\begin{equation}
x_{\rm front} = \left \lfloor -\frac{\arctan\frac{y}{x}} {R_{h}} \right \rfloor, 
y_{\rm front} = \left \lfloor -\frac{\arctan\frac{z}{\sqrt{x^2+y^2}}} {R_{v}} \right \rfloor
\end{equation}
where $R_{h}$ and $R_{v}$ are the horizontal and vertical angular resolution in radians, respectively. In order to project the label annotations onto the front view plane, we first calculated the bounding box in camera coordinates and transferred the corners to Velodyne coordinates by multiplying the transition matrix $T_{c2v}$. The object location was mapped onto the front view similarly, as illustrated in Fig. \ref{fig:kitti_process}. Based on the front view locations, objects were cropped with a fixed size to establish the recognition dataset. 

The changing from 3D to 2D would reduce the computational complexity of recognition. We also artificially generated arrival time $t_a$ for each event $(x, y, z)$ linearly with respect to $x$. The processed KITTI dataset contains 32456 training samples and 8000 testing samples covering 8 classes of KITTI objects. 

\paragraph{\bf N-Sydney} The N-Sydney Urban Objects dataset \cite{chen2014performance} is an event-based LiDAR dataset containing 26 object classes. We considered only the following 9 classes: Van, Tree, Building, Car, Truck, 4wd, Bus, Traffic light, and Pillar. We artificially generated arrival time $t_a$ for each event.

\subsection{DVS Datasets}

\paragraph{\bf DVS-CIFAR10}
The DVS-CIFAR10 dataset \cite{li2017cifar10} is converted from the popular CIFAR10 data set. It has 10000 samples covering 10 classes. We split the dataset into a training set of $8000$ samples and a testing set of $2000$ samples while adopting the full $128 \times 128$ pixel scene. 

\paragraph{\bf DVS-barrel}

The DVS-barrel dataset has 6753 samples with 36 classes \cite{orchard2015hfirst}, which we split into a training set of 3453 samples and a test set of 3000. We used the ``ExtractedStabilized'' version of the dataset, which, rather than using the full $128 \times 128$ pixel scene, extracts individual characters into $32 \times 32$ pixel scenes.

\paragraph{\bf N-MNIST, N-Caltech101, Hand-Gesture} The N-MNIST \cite{orchard2015hfirst} and N-Caltech101 \cite{zhang2006svm} datasets are the conversion of two popular image datasets MNIST and Caltech101. The N-MNIST has $10$ object classes and image size $28 \times 28$. Caltech101 has 100 object classes plus a background class. The image size on average is $200 \times 300$. The Hand-Gesture dataset \cite{huang2011gabor} is a DVS dataset with an image size of $120\times 320$.

\section{Evaluation}
\label{sec:evaluation}

\subsection{Experiment Setup}


Table \ref{tbl:snn_config} lists the network configurations we designed based on the proposed spiking learning system architecture for the datasets listed in Fig. \ref{fig:imagesamples}.

\begin{table}[t]
\centering
\caption{Our network models. Sample notation explained: F256 (fully connected layer with 256 spiking neurons), C5-48 (convolutional layer with 48 spiking kernels of size $5 \times 5$), AP (average-pooling layer with stride 2).}
\label{tbl:snn_config}
\scalebox{0.9}{
\begin{tabular}{r|l}
\hline
KITTI    & (50$\times$118$\times$1): C5-48, C5-24, F256, F8 \\ 
N-Sydney & (32$\times$32$\times$32): C5-32, C3-32, F128, F9 \\ \hline \hline

DVS-barrel & (input 1024): F2000, F36 \\
N-MNIST & (28$\times$28$\times$1): C5-32, C5-16, F10 \\
N-Caltech101 & (200$\times$300): C5-16, C3-8, F64, F101 \\
HandGesture & (120$\times$320): C5-32,C3-48,C3-16,F64,F10 \\ \hline

    & Small: C3-32, AP, C3-48, AP, F256, F10 \\
 DVS- & Medium: C3-32, C3-48, AP, C3-64, AP, \\
  CIFAR10         &  \hspace{1.5cm} F256, F10 \\
 (128$\times$128)            & Large: C3-32, C3-64, AP, C3-128, \\
            & \hspace{1cm} C3-256, AP, F1024, F10 \\ \hline
\end{tabular}}
\end{table}


All datasets were tested over models with multiple SCNN and FC layers because our experiments showed that they were much better than simpler SNNs with FC layers only. 
As for the KITTI dataset, a model with two SCNN layers and two FC layers was employed. The input size was $50\times118\times1$. The kernel size for the SCNN layers was $5\times5$, with a stride size of 2. The numbers of kernels were 48 and 24, respectively. The output from the second SCNN layer had a size $13\times30\times24$ and was flattened and passed to the first FC layer (with 256 spiking neurons). The second FC layer had 8 output channels. The batch size was set to 10, and the initial learning rate was 1e-3 with decay. Adam optimizer was adopted for training.

The N-Sydney Urban Object dataset and the N-Caltech 101 dataset were tested over similar models with two SCNN layers and two FC layers. The Hand-Gesture dataset was tested over a model with three SCNN layers and two FC layers.

The DVS-CIFAR10 dataset was considered the most challenging one among these datasets. It is also much more challenging than the conventional frame-based CIFAR10 due to noisy samples and a single intensity channel. We created three different spiking network structures for a fair comparison with \cite{wu2019direct}, which also created three SNN structures to compare with other SNN and DNN results and published the best accuracy so far. The training employed Adam as the optimizer with a batch size of 8, and 100 training epochs with an exponentially decaying learning rate. By manipulating learning rates, the fastest convergence was obtained when the learning rate started at 1e-2 in epoch 1 and ended at 1e-5 in epoch 100.

Note that we tried various optimizers such as SGD and Adam on each model during training. Their training performance did not show a significant difference.

\subsection{Experiment Results}
We used recognition accuracy ($A$) and recognition (inference) delay ($D$) as performance metrics to evaluate and compare our models with the state of the arts. Based on $A$ and $D$ we calculated performance gain ($G$) of our model as
\begin{equation}
    G_{\rm acc} = \frac{A_{\rm ours}-A_{\rm ref}}{A_{\rm ref}}, \;\;\;
    G_{\rm time} = \frac{D_{\rm ref} - D_{\rm ours}}{D_{\rm ref}},
\end{equation}
where $G_{\rm acc}$ and $G_{\rm time}$ are the accuracy gain and time efficiency gain of our model over some reference model, respectively. The time efficiency gain is also the ratio of delay/latency reduction. The delay includes both the delay of the inference algorithm, which we call ``inference delay", and the delay caused by waiting for the asynchronous arrival of events. Their sum is the ``total delay". Although we do not directly evaluate computational complex and energy efficiency, we must point out that SNN-based models, in general, have lower computational complexity and higher energy efficiency, as pointed out in \cite{zhou2020deepscnn} and many other SNN publications.

\paragraph{\bf KITTI}
We take the transformed KITTI dataset as a representative of LiDAR datasets to interpret the evaluation results in detail. The results of other datasets are provided at the end of this section. For the KITTI dataset, we compared our proposed system against the VGG-16 model of conventional CNN. To make the processed KITTI dataset work on VGG-16, we replicated the single intensity channel into three RGB color channels and resized the image from $50\times118$ to $128 \times 128$. We utilized the VGG-16 pre-trained on ImageNet for transfer learning. Table \ref{tbl:acc_kitti} shows that our system not only achieved better accuracy (with a gain of $G_{\rm acc}=5.46\%$ over VGG-16), but also had much smaller latency or higher time efficiency (with a gain of $G_{\rm time}$ between $56.3\%$ and $91.7\%$ over VGG-16). The reason for VGG-16 to have relatively lower testing accuracy might be because the processed KITTI dataset had a single intensity channel and had a smaller image size than the ideal VGG-16 inputs. A smaller network quite often can achieve better accuracy performance than a complex network over a relatively small dataset.

\begin{table}[t]
\caption{Comparison of our SNN model with the VGG-16 model over the KITTI dataset for accuracy and timing efficiency.}
\label{tbl:acc_kitti}
\begin{center}
\scalebox{0.9}{
\begin{tabular}{cc c c} 
\hline
Model   &  VGG-16   & Our Model  & Gain \\ \hline
Accuracy &  91.6\%   & 96.6\%      &   5.46\%   \\ \hline
Inf. Delay (CPU) &  38 ms  & 8.5 ms  &  77.6\%   \\
Total Delay (CPU) & 63 ms  & 27 ms   &  57.1\%  \\ \hline
Inf. Delay (GPU) & 23 ms & 1.9 ms    &  91.7\%   \\
Total Delay (GPU) & 48 ms & 21 ms  &   56.3\%   \\ \hline
\end{tabular}}
\end{center}
\end{table}

Next, let us focus on the delay and time efficiency comparison results. 
To obtain an inference delay, we set temporal coding parameter $\beta$ for our system to work in synchronous mode, which gave us the SNN's average running time for an image. We did this over both Intel Core i9 CPU and NVIDIA GeForce RTX 2080Ti GPU. On GPU, our model spent 1.9 ms (millisecond) while VGG-16 needed 23 ms. Our model ran faster because our model both was much simpler and ran asynchronously (inference may terminate at a very small short spiking time $t_{out}$ at the last layer). 
To obtain the total delay, we set a temporal coding parameter to work in the asynchronous mode to exploit both the asynchronous event arrival property and the SNN's asynchronous processing property. By contrast, VGG-16 had to wait until all the events were received before starting processing. Since KITTI LiDAR generated $360^\circ$ image frames at the speed of $10$ Hz, the collection of all events for a $90^\circ$ field-of-view image had a delay of $25$ ms. So the total delay on GPU was $25+23=48$ ms. In contrast, our model had a total delay of $21$ ms on average, a delay reduction (and time efficiency gain) of $56.3\%$. We can also see that the asynchronous events arrival time dominated the total delay. Our model just needed a fraction of events for inference, so it had a much smaller delay. 

\begin{figure}[t]
    \centering
    \includegraphics[width=\linewidth]{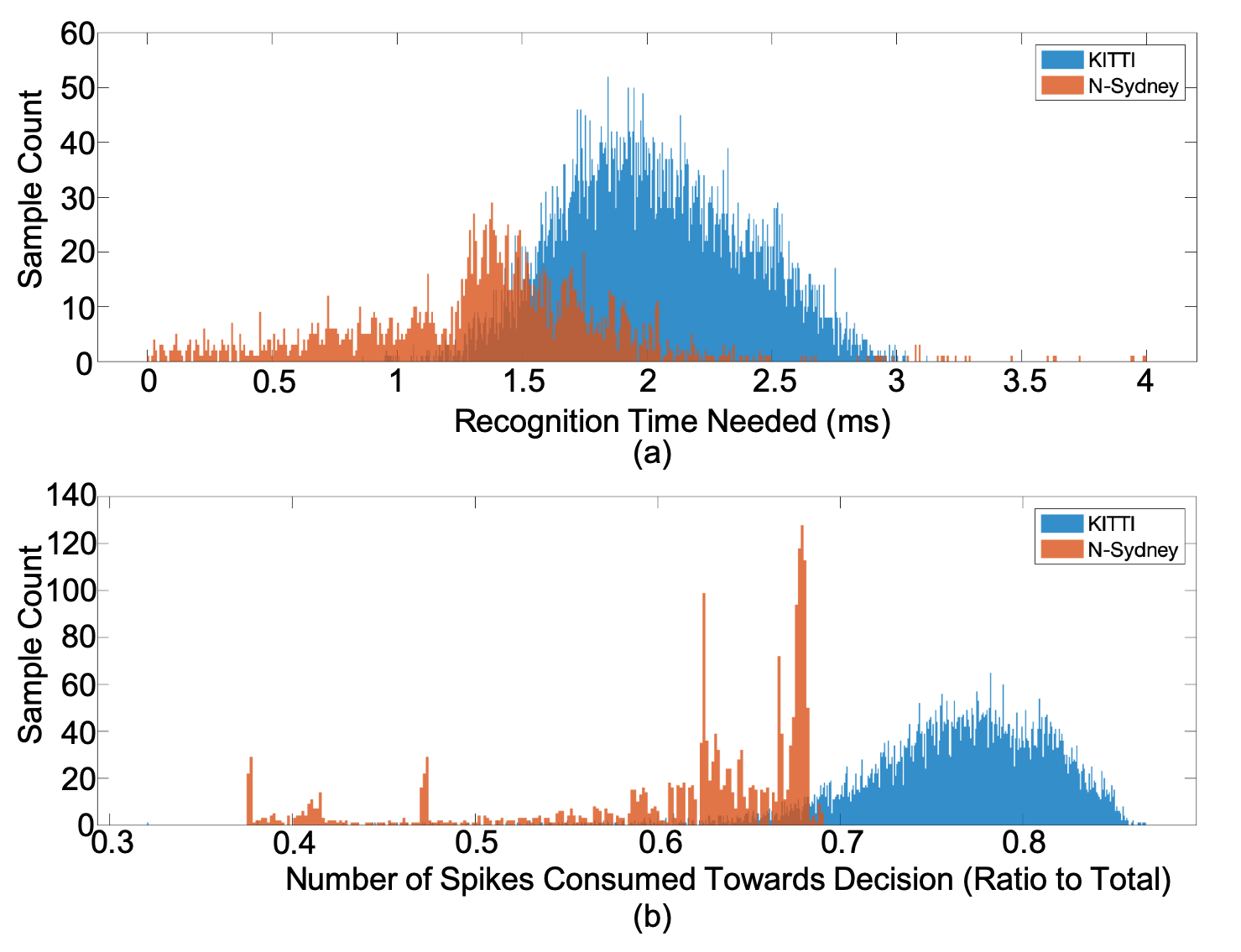}
    \caption{Distributions of (a) inference time and (b) ratio of events used in inference, on LiDAR datasets.}
    \label{fig:effi_lidar}
\end{figure}

\begin{figure}[t]
    \centering
    \includegraphics[width=\linewidth]{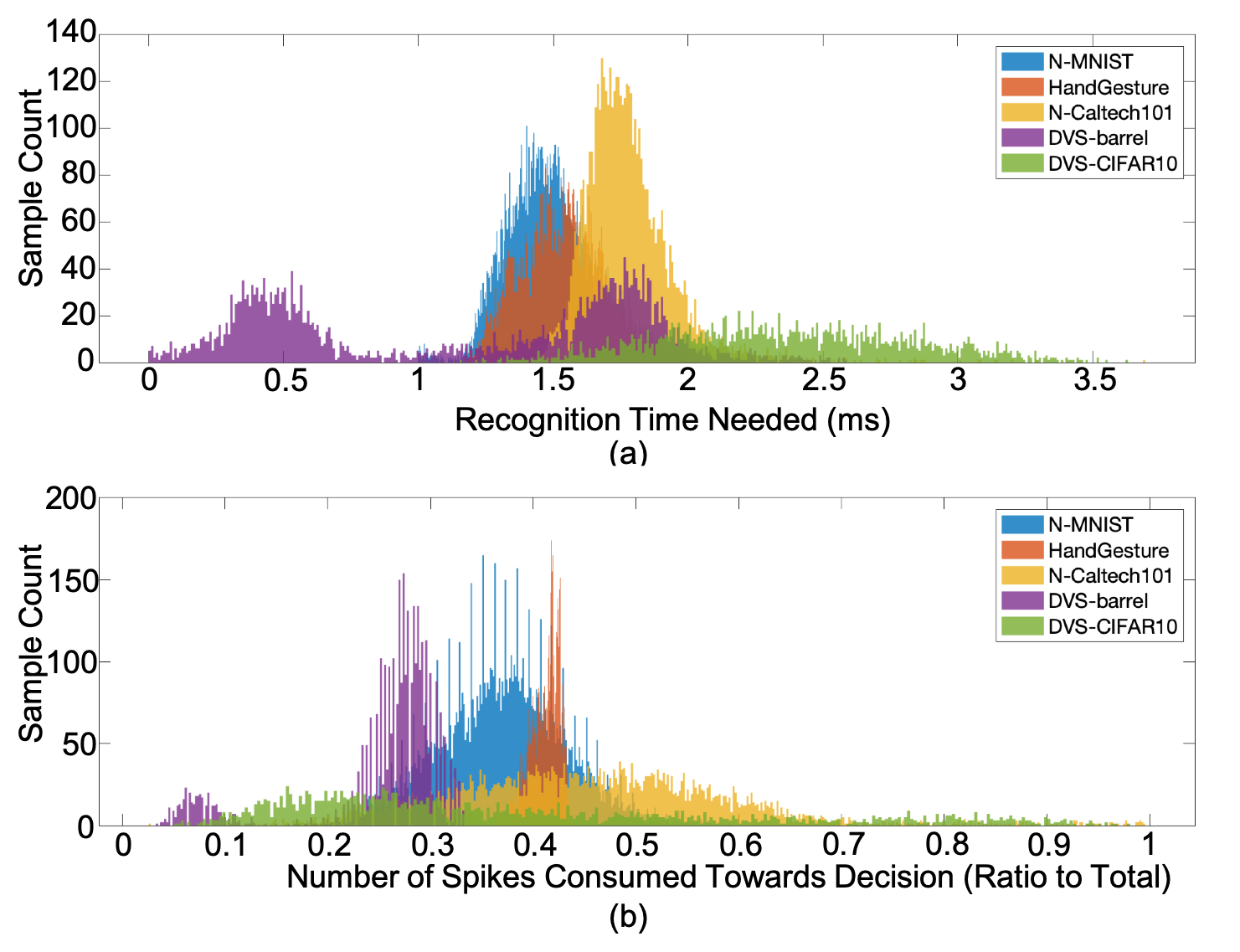}
    \caption{Distributions of (a) inference time and (b) ratio of events used in inference, on DVS datasets.}
    \label{fig:effi_dvs}
\end{figure}

The distribution of the ratio of events used by our model on the KITTI dataset is shown in Fig. \ref{fig:effi_lidar}(b). On average, $76\%$ of events were used in inference. These figures also demonstrate that our system worked asynchronously and selected automatically various numbers of events in different recognition tasks. In addition, we calculated the ``ideal inference delay", which is defined as the difference between the SNN output spike time and the first SNN input spike time. Obviously, ``ideal" means that we skip hardware processing delay. The distribution is shown in Fig. \ref{fig:effi_lidar}(a). An interesting observation was that the ``ideal inference time" was approximately $1.9$ ms, the same as what we obtained from practical GPU running time in Table \ref{tbl:acc_kitti}. This may not be surprising because the SNN spiking time duration is usually much longer than GPU executing duration for simple networks. Asynchronous SNN operation can surely reduce recognition/inference delay.

Based on the above interesting observation, we propose event-ratio as an approximation of time efficiency gain. The event-ratio is defined as the proportion of contributing events (input events consumed before the decision) for recognizing the object in an image frame to all the events corresponding to this image frame, i.e.,

\begin{equation}
r_{\rm event} = \frac{N_{\rm contributing}}{N_{\rm all}}.
\end{equation}

The estimated time efficiency gain is defined as
\begin{equation} 
\hat{G}_{\rm time} \approx  1 - r_{\rm event}.  \label{eq4.1}
\end{equation}

The estimation is accurate when the computation delay of CPU and GPU is negligible compared with the event accumulation time duration, which is often true. 
This way of calculating the time efficiency gain resolves a big hurdle in DVS datasets because DVS datasets usually do not have realistic events timing. Even for LiDAR datasets, event timing could only be constructed artificially. The time efficiency gain (\ref{eq4.1}) catches the key factor, i.e., asynchronous operation, and skips the variations caused by non-ideal software coding and platforms. 

Adopting this approximation, the accuracy and time efficiency of our models on all the datasets are listed in Table \ref{tbl:acc_delay}. Gain $G_{\rm acc}$ was calculated by choosing the accuracy of the best model used in our comparison as $G_{\rm ref}$, see Table \ref{tbl:comp_cifar10} and Table \ref{tbl:acc_sydney}. From the table, we can see that our system, in general, was competitive with the state-of-the-art models in recognition accuracy. More importantly, our system needed only a fraction of the events, which lead to a $34\%$ to $75\%$ gain in time efficiency. 

\begin{table}[t]
\caption{Summary of accuracy and time efficiency of our system.}
\label{tbl:acc_delay}
\begin{center}
\scalebox{1}{
\begin{tabular}{c | c c c c} 
\hline
Dataset & Accuracy & $G_{\rm acc}$      &  Event & $\hat{G}_{\rm time}$  \\ 
        &        &          & Ratio         &    \\ \hline
KITTI      & 96.62\% &  5.46\%       & 0.76        &  24\% \\ 
N-Sydney   &  78.00\% & 6.85\%  &  0.62        &  38\% \\ \hline
DVS-CIFAR10 & 69.87\%  & 8.71\%  &  0.38   & 62\% \\
DVS-Barrel & 99.52\% &  4.32\%  &  0.25  & 75\%  \\
N-MNIST  & 99.19\%  &   -0.0\%  &  0.37  &  63\% \\
N-Caltech101 & 91.89\% & -1.6\% & 0.45  & 55\% \\
Hand-Gesture & 99.99\% & 0.91\% &  0.41  & 59\% \\ \hline
\end{tabular}}
\end{center}
\end{table}


\paragraph{\bf DVS-CIFAR10}
We take DVS-CIFAR10 as an example to detail the evaluations on DVS datasets. The results of all other DVS datasets are given at the end of this section. Table \ref{tbl:comp_cifar10} shows that our model had 69.87\%\ recognition accuracy, higher than competitive models listed. 
Note that the competitive models were selected carefully according to their importance for the development of this dataset and their state-of-art performance. Their accuracy values were cited directly from the papers. Their methods were also listed to show clearly the performance difference of conventional machine learning methods, DNN/CNN, and SNN/SCNN.
The reduction of the delay was even more striking based on the description of Table \ref{tbl:acc_delay}. From Fig. \ref{fig:effi_dvs}(b), we can see that our model used only $38\%$ of events for inference, which means a delay reduction of over $62\%$. From Fig. \ref{fig:effi_dvs}(a), our model on average used $2.35$ ms for inference based on our artificially created timing information of the events similar to KITTI. 

\begin{table}[t]
\centering
\caption{Comparison with existing results on DVS-CIFAR10}
\label{tbl:comp_cifar10}
\scalebox{0.9}{
\begin{tabular}{cccc}
\hline
Model      & Method     & Accuracy  \\ \hline
Zhao 2014 \cite{zhao2014feedforward}    & SNN                & 22.1\%            \\
Lagorce 2016 \cite{lagorce2016hots} & HOTS                & 27.1\%            \\
Sironi 2018 \cite{sironi2018hats}  & HAT                 & 52.4\%            \\
Wu 2019 \cite{wu2019direct}      & SNN   & 60.5\%            \\
Our model (sml)     & SCNN                & 60.8\%            \\
Our model (mid)     & SCNN                & 64.3\%            \\
Our model (large)   & SCNN                & 69.9\%          \\ \hline
\end{tabular}}
\end{table}

\begin{figure}[t]
    \centering
    \includegraphics[width=0.9\linewidth]{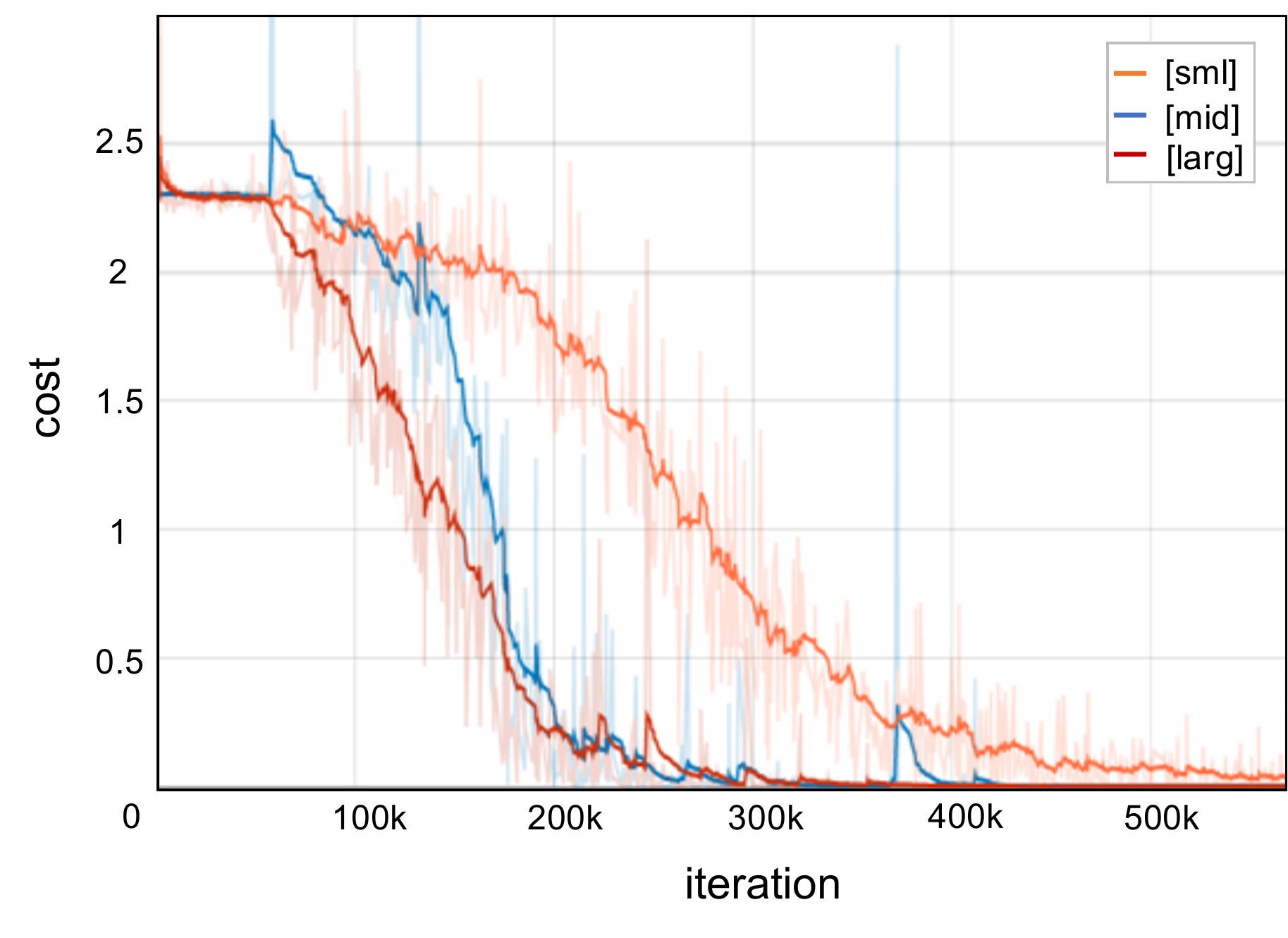}
    \caption{Training convergence of the proposed models over the DVS-CIFAR10 dataset.}
    \label{fig:loss_trend}
\end{figure}

The effect of different network configurations on the training process was investigated and depicted in Fig. \ref{fig:loss_trend}, where the training loss was calculated with 8 images randomly sampled from the training set. When being trained on the DVS-CIFAR10 dataset, the larger model converged faster than the smaller one, which might be because of the better capability of larger SNN in learning data representations.  

\paragraph{\bf Other datasets}
For the rest of the datasets, N-Sydney, DVS-barrel, N-MNIST, N-Caltech101, and Hand-Gesture, the accuracy comparisons with the state-of-the-art models are listed in Table \ref{tbl:acc_sydney}. The table also listed the major network architecture. We can see that our models provided competitive recognition accuracy. For challenging datasets with relatively lower recognition accuracy, such as N-Sydney, our model had much higher performance. For the popular N-MNIST dataset, our model had relatively low computational complexity because the number of trainable weights (2.2e4) was lower than other models. 

The event-ratio/inference-time distributions of our SNN model on LiDAR and DVS datasets are shown respectively in Fig. \ref{fig:effi_lidar} and Fig. \ref{fig:effi_dvs}. On average, $62\%$ of events were needed for the N-Sydney dataset, while the DVS datasets mostly took less than $50\%$ of events to accomplish the inference. It can also be seen that the inference time for most samples was in the range of $1$ to $3$ ms, demonstrating a strong performance in time efficiency.

\begin{table*}[t]
\caption{Performance comparison over N-Sydney, N-MNIST, N-Caltech101 and Hand-Gesture datasets.}
\label{tbl:acc_sydney}
\begin{center}
\scalebox{1}{
\begin{tabular}{c | c c c} 
\hline
Dataset &  Model  & Accuracy &  Method \\ \hline \hline
  N-&  Chen'14 \cite{chen2014performance}     &  71\%  & GFH+SVM  \\
Sydney  & Maturana'15 \cite{maturana2015voxnet} &  73\%  & CNN \\
          & Our Model      & 78\%   & SCNN \\ \hline \hline

   & Perez'13 \cite{perez2013mapping}   & 95.2\% & CNN          \\
 DVS-     & Perez'13 \cite{perez2013mapping}    & 91.6\% & SCNN           \\
Barrel     & Orchard'15 \cite{orchard2015hfirst}  & 84.9\% & HFirst          \\
     & Our Model    & 99.5\%    & SCNN        \\ \hline \hline

        & Orchard'15 \cite{orchard2015hfirst} & 71.20\% & HFisrt, 1.2e5 \\ 
        & Neil'16 \cite{neil2016phased} & 97.30\% & LSTM, 4.5e4 \\
N-         & Lee'16 \cite{lee2016training} & 98.70\% & SNN, 1.8e6 \\
MNIST        & Shreshtha'18 \cite{shrestha2018slayer} & 99.20\%  & SNN, 7.2e4 \\
        & Wu'18 \cite{wu2018spatio} & 98.78\% & SNN, 1.8e6 \\
        & Our Model & 99.15\% & SCNN, 2.2e4 \\ \hline \hline

   & Zhang'06 \cite{zhang2006svm} & 66.23\% & SVM-KNN \\
   & Donahue'14 \cite{donahue2014decaf} & 86.91\% & DNN \\
  N- & Chatfield'14 \cite{chatfield2014return} & 88.54\% & CNN \\
 Caltech  & He'15 \cite{he2015spatial} & 93.42\% & CNN \\
101 & Orchard'15 \cite{orchard2015hfirst} & 5.4\% & HFirst, SNN \\        
   & Sironi'18 \cite{sironi2018hats} & 64.2\% & HATS, SNN \\
    & Our Model & 91.9\% & SCNN \\ \hline \hline
 
      & Huang'11 \cite{huang2011gabor} & 94\% & GF \\
 Hand- & Mantecon'16 \cite{mantecon2016hand} & 99\% & SVM \\
 Gesture & Our Model & 99.9\% & SCNN \\ \hline
\end{tabular}}
\end{center}
\end{table*}

\section{Conclusion and Future Work}
\label{sec:conclusion}

In this paper, we proposed a spiking learning system utilizing temporal-coded spike neural networks for efficient object recognition from event-based sensors such as LiDAR and DVS. A novel temporal coding scheme was developed, which permits the system to exploit the asynchronously arrived sensor events without delay. Integrating nicely with the asynchronous processing nature of SNN, the system can achieve superior timing efficiency. The performance of the system was evaluated on a list of $7$ LiDAR and DVS datasets. The experiment proved that the proposed method achieved remarkable accuracy on real-world data and significantly reduced recognition delay. This paper demonstrates the potential of SNN in challenging applications involving high speed and dynamics.

On the other hand, although we developed the general temporal encoding scheme in Section~\ref{sec:snn_model}, the hyper-parameters of the encoding rules (\ref{eq3.25}) (\ref{eq3.26}) were not optimized in our experiments. But rather, we used some hand-picked parameter values heuristically. Thanks to the cost function (\ref{eq3.50}), the training of SNN led to the minimum spike timing $t_{L,i}$ in the final output layer, which means that we still obtained a self-optimized event accumulation time $t$. It will be an interesting future research topic to optimize directly these hyper-parameters by either including them into the cost function (\ref{eq3.50}) or using the genetic algorithm. This will optimize the event accumulation time $t$ to further enhance recognition accuracy and reduce recognition delay. In addition, we evaluated our learning system for single-object recognition or classification in every single image only. It remains to extend this system to object detection and/or continuous video processing, which will make the study of time efficiency more interesting.  Considering the small size of the event-driven datasets used in this paper, only relatively simple SNNs were applied because there would be over-fitting otherwise. This might be one of the reasons for our SNN model's surprisingly better performance than VGG-16 shown in Table \ref{tbl:acc_kitti}. It will be an interesting work to adapt our system to large event-based datasets in the future.
   
\newpage
\bibliographystyle{named}
\bibliography{ijcai20}

\end{document}